\documentclass[letterpaper]{article} 
\pdfoutput=1
\usepackage{aaai24}  
\usepackage{times}  
\usepackage{helvet}  
\usepackage{courier}  
\usepackage[hyphens]{url}  
\usepackage{graphicx} 
\usepackage{natbib}  
\usepackage{caption} 
\usepackage{booktabs}
\usepackage{multirow}
\usepackage{amsmath}
\usepackage{amssymb}
\usepackage{algorithm}
\usepackage{algorithmic}
\usepackage{makecell}

\usepackage{newfloat}
\usepackage{listings}
\DeclareCaptionStyle{ruled}{labelfont=normalfont,labelsep=colon,strut=off} 
\floatstyle{ruled}
\newfloat{listing}{tb}{lst}{}
\floatname{listing}{Listing}
\title{Decompose Semantic Shifts for Composed Image Retrieval}
\author{
    Xingyu Yang\textsuperscript{\rm 1}\thanks{Contribution during internship at JD Explore Academy.},
    Daqing Liu\textsuperscript{\rm 2},
    Heng Zhang\textsuperscript{\rm 3},
    Yong Luo\textsuperscript{\rm 3}\thanks{Corresponding author.},
    Chaoyue Wang\textsuperscript{\rm 3},
    Jing Zhang\textsuperscript{\rm 3}
}
\affiliations{
    \textsuperscript{\rm 1}School of Computer Science and National Engineering Research Center for Multimedia Software, Wuhan University\\
    \textsuperscript{\rm 2}JD Explore Academy,
    \textsuperscript{\rm 3}Renmin University of China,
    \textsuperscript{\rm 4}The University of Sydney\\
    \{yangxingyu2021,luoyong\}@whu.edu.cn, \{liudq.ustc,jingzhang.cv\}@gmail.com, \\ \{zhangheng\}@ruc.edu.cn, \{chaoyue.wang\}@outlook.com
}

\usepackage{bibentry}

\begin{document}

\maketitle

\begin{abstract}
Composed image retrieval is a type of image retrieval task where the user provides a reference image as a starting point and specifies a text on how to shift from the starting point to the desired target image.
However, most existing methods focus on the composition learning of text and reference images and oversimplify the text as a description, neglecting the inherent structure and the user's shifting intention of the texts. As a result, these methods typically take shortcuts that disregard the visual cue of the reference images.
To address this issue, we reconsider the text as instructions and propose a Semantic Shift network (SSN) that explicitly decomposes the semantic shifts into two steps: from the reference image to the visual prototype and from the visual prototype to the target image.
Specifically, SSN explicitly decomposes the instructions into two components: degradation and upgradation, where the degradation is used to picture the visual prototype from the reference image, while the upgradation is used to enrich the visual prototype into the final representations to retrieve the desired target image.
The experimental results show that the proposed SSN demonstrates a significant improvement of 5.42\% and 1.37\% on the CIRR and FashionIQ datasets, respectively, and establishes a new state-of-the-art performance. Codes will be publicly available.
\end{abstract}

\begin{figure}[t]
\begin{center}
\includegraphics[width=\linewidth]{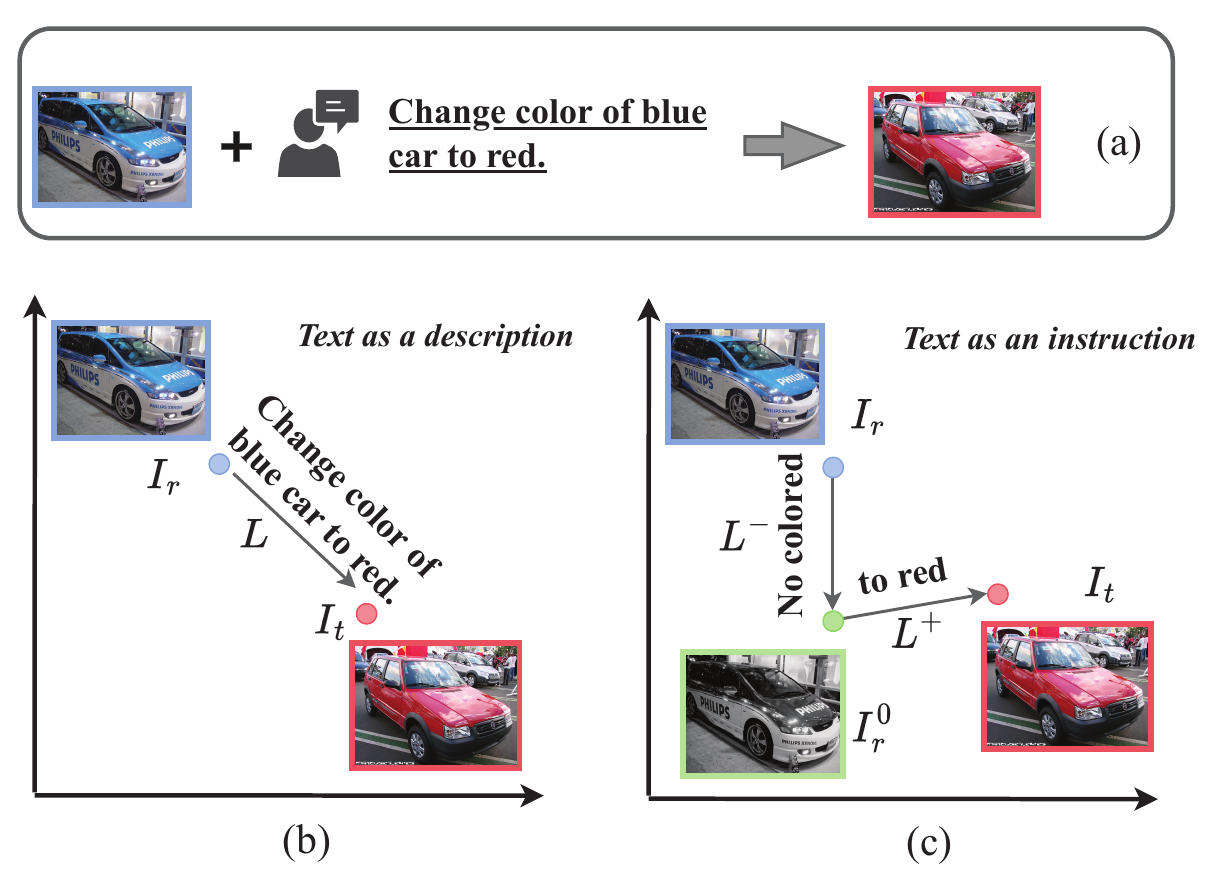}
\end{center}
   \caption{CIR task aims to retrieve the target image according to a pair of reference image and modification text provided by the user. \textbf{Bottom Left}: (a) Existing works treat text as a description that connects the reference image and the target image. They follow the paradigm of $I_{r}+L \leftrightarrow I_{t}$. \textbf{Bottom Right}: (b) We propose to consider the text as an instruction, inheriting the property of human language to express semantic shifts. With text instructions, the reference image is first degraded into visual prototypes and then enriched into the final representations to retrieve. This process can be described as $I_{r} \stackrel{L^-}{\longrightarrow} I_{r}^0 \stackrel{L^+}{\longrightarrow} I_{t}$. }
\label{fig:comparision}
\end{figure}
\section{Introduction}\label{sec:intro}
Composed Image Retrieval~\cite{vo2019composing} (CIR) is an emerging image retrieval task in that the users can provide a multi-modal query composed of a reference image and a text. Different from the traditional image retrieval~\cite{weinzaepfel2022learning} where the users must provide the exact same image of the desired result or text-to-image retrieval~\cite{wang2019camp} where the users should describe the target in a detailed language, as shown in Figure~\ref{fig:comparision}a, CIR relax the requirement of input thus the users can simply provide an example image that similar to the desired image as reference, and then describe the difference from the reference to the target.
Despite their diverse model architectures~\cite{kim2021dual,DBLP:conf/mm/Yang0ZL21}, the essence of this task is to fully understand the user's intent conveyed by the reference image and the language and then find the most similar image from all candidates.
Thanks to the development of vision features~\cite{radford2021learning} and language representations~\cite{devlin2018bert}, we can accomplish the CIR task with more flexible free-form languages, including changing one specific attribute of one object and adding or removing some objects.

However, capturing the user's intentions is still a challenging problem because the instruction text is far different from the description text that is commonly used in current vision-language tasks, eg., visual grounding~\cite{deng2018visual}, cross-modal retrieval~\cite{wang2019camp}, or visual captioning~\cite{liu2022show}.
For example, given the reference image and the text ``Changle the color of the blue car into red'' in Figure~\ref{fig:comparision}b, how to depict the desired image for us humans? One may have the following intuitive procedure: 1) identify the part of the reference image that should change, ie, ``the color of the blue car'', 2) imagine the visual prototype of the reference image, ie, this car without any color, and 3) picture the final desired target image as ``the car in red''.

Unfortunately, despite the complex model architecture design with cross-modal attention~\cite{hosseinzadeh2020composed}, graph neural network~\cite{DBLP:conf/mm/ZhangYZX22}, or fine-grained visual network~\cite{DBLP:conf/cvpr/HosseinzadehW20}, existing methods~\cite{baldrati2022conditioned,zhao2022progressive,DBLP:conf/cvpr/GoenkaZJC0HN22} generally oversimplify the CIR task as a composition learning of vision and language where the text is usually treated as a description (Figure~\ref{fig:comparision}), disregarding the propriety of the structure of the text, which should be an instruction on how to modify the reference to the target.
More seriously, composition learning typically introduces redundant or even incorrect information that may disrupt the final representations, eg, simply combining the semantics `blue, car' of the reference image and the semantics `red, car, blue' of the text will depict `a red and blue car' flawedly. The fundamental cause of this issue lies in a lack of precise understanding of the language.

In this paper, we propose to take the text as instructions that represent the semantic shifts from the reference image to the target image, and then decompose the instructions into two parts: the degradation and the upgradation.
As illustrated in Figure~\ref{fig:comparision}c, based on the decomposition, we conduct the desired image representations in two steps: 1) degrading the reference image into the visual prototype that only contains the visual attributes that need to be preserved, and 2) upgrading the visual prototype into the final desired target.
Thanks to the decomposition of instructions, we divide the complex task that models the user's intentions into two simple and orthogonal sub-tasks which are easier to learn. Based on the final representations, we can directly find the nearest neighbors in the latent space as the final retrieval results.

Specifically, we implement the proposed method with a Semantic Shift Network (SSN) that is composed of four components: 1) the representation networks that extract visual and language features of reference images, target images, and instructions; 2) the decomposing network that decomposes the instruction text into the degradation part and upgradation part; 3) the degrading network that transforms the reference image to the visual prototype conditioned on the degradation part of the instruction text; 4) the upgrading network that transforms the visual prototype to the final representation of desired image conditioned on the upgradation part. To train the SSN, we design a traditional retrieval loss to guarantee the overall performance of composed image retrieval, as well as a regularization constraint that disciplines the language decomposing and the visual prototypes.

We validate the effectiveness of SSN on two widely-used composed image retrieval benchmarks, ie, FashionIQ~\cite{wu2021fashion} and CIRR~\cite{DBLP:conf/iccv/0002OTG21}. SSN stands as a new state-of-the-art on all metrics. Specifically, we achieve impressive improvements of 5.42\% and 1.37\% on CIRR and FashionIQ mean recall metrics, respectively.

In summary, our contributions include:
\begin{itemize}
	\item We reformulate the composed image retrieval task as a semantic shift problem based on the text instructions, with the shift path as reference image $\rightarrow$ visual prototype $\rightarrow$ desired target image.

	\item We introduce a Semantic Shift Network for the CIR task that implements the decomposed semantic shifts with several well-designed components.

	\item The proposed SSN achieves state-of-the-art performance with impressive improvements on two widely-used composed image retrieval datasets.
\end{itemize}
\begin{figure*}
\begin{center}
\includegraphics[width=\linewidth]{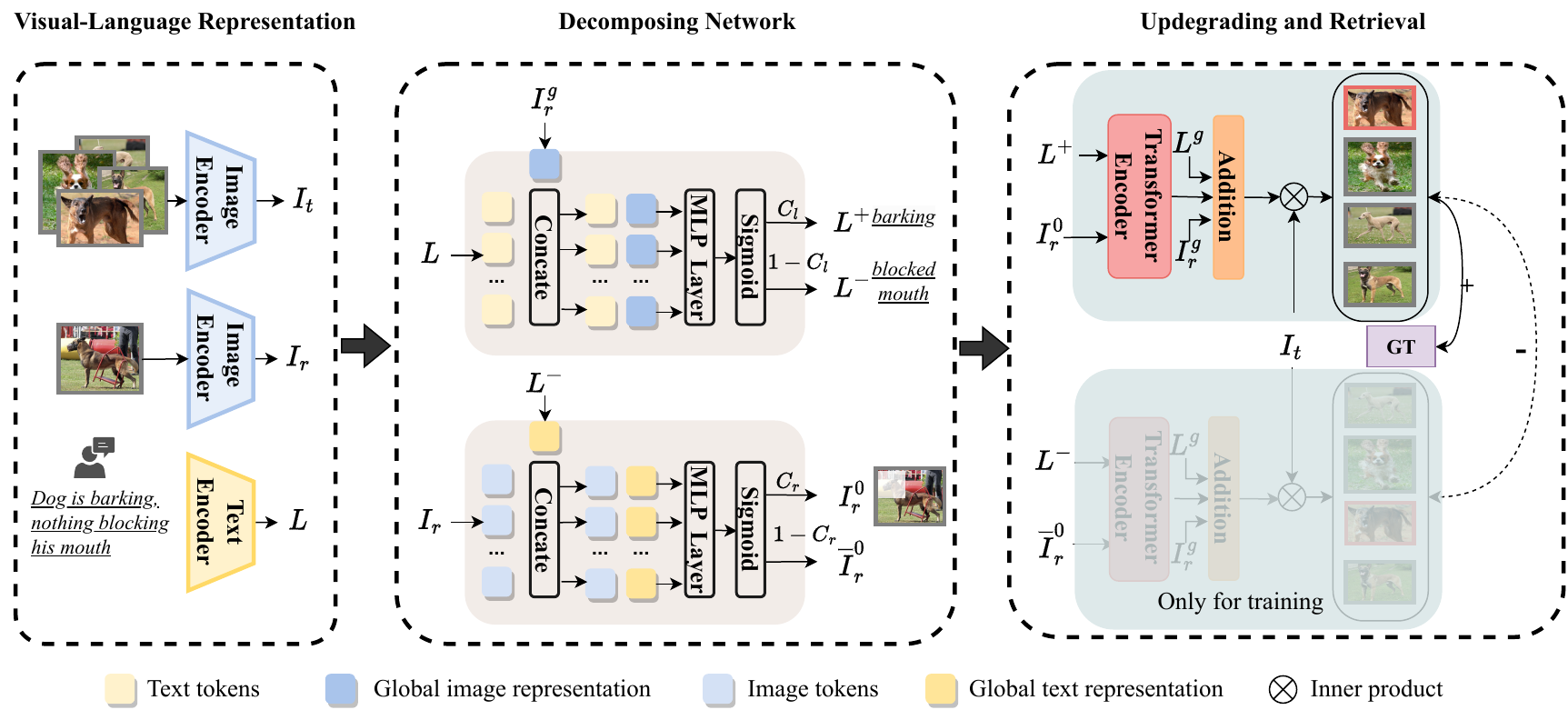}
\end{center}
   \caption{The pipeline of our proposed Semantic Shift Network (SSN). Given a pair of reference images and text modifiers (also as an instruction), we aim at retrieving the correct target image from candidate images. At the stage of visual-language representation, we utilize the CLIP image and text encoders to obtain the respective features. Then the semantic shift features from text instruction are decomposed to direct the reference image features into a visual prototype $I_r^0$. The text features biased toward the target and reference image (namely $L^+$ and $L^-$ respectively) are generated at the same time. In the late composite module, $L^+$ is fused with visual prototypes $I_r^0$ by transformer a encode layer and then are linearly added to global representations. Finally, similarity scores are measured by an inner-product operation to generate the ranked list. }
\label{fig:overview}
\end{figure*}
\section{Related Work}
In this section, we first briefly background for conventional image retrieval and then present works of composed image retrieval, which is an extension of image retrieval but with a multi-modal query.

\noindent\textbf{Image Retrieval}.
Image Retrieval is a fundamental task for the computer vision community since it has a wide range of application scenarios, eg, search engines, and e-commerce. Given a query image, we need to return the most similar image stored in a large-scale database. In the beginning, global image representation~\cite{chen2022deep} based retrieval methods were investigated. In order to achieve fine-grained matching~\cite{sun2021loftr} between images and thus improve retrieval performance, several approaches transformed images into a number of local representations (eg, region features). However, these pioneering works~\cite{chen2022deep,sun2021loftr,weinzaepfel2022learning}' queries are images only and they focus more on similarity matching between images. In reality, people usually convey query intent with text rather than images, so text-image retrieval is also a research focus in image search. Benefiting from the success of a large model for visual language pre-training~\cite{kim2021vilt,lu2019vilbert}, cross-modal representations have shown remarkable performance in text-image retrieval tasks. Moreover, the query can also be a combination of image and text, which is the direction we explore in this work.

\noindent\textbf{Composed Image Retrieval}.
Composed Image Retrieval refers to searching target images given semantically related reference images and modification texts. The text contains specified changes guiding the reference image closer to the target image. Most existing CIR models~\cite{Goenka2022} are applied in the field of fashion domain rather than the open domain. One line of work introduces target images into the forward process, which greatly increases the training cost and is difficult to apply in practice. Another line of works aiming at efficient retrieval explores composition learning and our model belongs to this type. Gated residual fusion is first proposed to combine image and text features for the CIR task in TIRG~\cite{vo2019composing} and is commonly used for global fusion in later works~\cite{wang2022exploring,kim2021dual,Chen2020LearningJV}. MAAF~\cite{dodds2020modality} method applies a self-attention mechanism to concatenated image and text sequences. Conditioned on text, VAL~\cite{chen2020image} is proposed to obtain combined features through multi-grained cross-modal semantic alignment, and CosMo~\cite{lee2021cosmo} is used to refine reference image features from the term of style and content. Benefiting from the visual-language models pre-trained on large-scale image text datasets, several works have applied them to the CIR downstream task and achieved favorable performance, eg, CIRPLANT~\cite{DBLP:conf/iccv/0002OTG21} and FashionVLIP~\cite{DBLP:conf/cvpr/fashionvlp}. Recently, a CLIP-based model equipped with a simple combiner network has shown great potential. Following this work, we also use the same CLIP encoder. Compared to previous works, our model didn't consider the modification text as a description, but as an instruction. This guides the reference image to semantically shift and back to a visual prototype (for a pair of reference and target images).
\section{Approach}
In this section, we present the proposed Semantic Shift Network (SSN) for composed image retrieval. In the following, we first briefly describe how to obtain the representations for image and text inputs in Section~\ref{sec:representation}. Then we present the technical details of the two steps for degradation and upgradation in Section~\ref{sec:degradation} and Section~\ref{sec:upgradation}, respectively, and finally depict our training objectives in Section~\ref{sec:trainandinfer}.
\subsection{Preliminaries}
The Composed Image Retrieval (CIR) task replaces the query in traditional image retrieval with multi-modal input, usually an image plus a text modifier. In the CIR setting, the query image is referred to as a reference image $r$, and the text modifier is denoted as $l$. Given each query $q=(r,l)$, the trained model returns a ranked list of the candidate images from a large image gallery $\mathcal{D}$, in descending order of similarity to the joint query semantic representation. An ideal retrieval system should rank the ground-truth target image $t$ at the first position.

We propose to model the composed image retrieval as a degradation-upgradation learning process where we treat the text as an instruction. It is described as: $I_{r} \stackrel{L^-}{\longrightarrow} I_{r}^0 \stackrel{L^+}{\longrightarrow} I_{t}$. By decomposing semantic shifts, the text instruction first directs the reference image toward the shared characteristics (defined as visual prototypes $I_r^0$) to retain essential visual cues. On the basis of this, $L^+$ further guides the visual prototypes $I_r^0$ close to the target image features $I_t$ by compositional learning. Furthermore, the two-step strategy reduces the difficulty of compositional learning due to fewer interruptions.
\subsection{Visual-Language Representation}
\label{sec:representation}
CLIP~\cite{radford2021learning} is a recently successful visual-language pre-trained model learned contrastively from 400M associated image-text pairs crawled from the internet. Multimodal representations extracted by the CLIP encoder can provide a good starting point and generalize well to a variety of downstream tasks. 

To leverage the powerful representation capability of CLIP, we adopt CLIP matched encoder to yield image features and text features. Formally, we denote the CLIP image encoder as $\varPhi_I$ and the CLIP text encoder as $\varPhi_L$. Given a triplet $(r,l,t)$ of reference image $r$, modification text $l$, and target image $t$, image features $I_{r}=\varPhi_I(r),I_{t}=\varPhi_I(t)$ and text feature $L$ is represented as $\varPhi_L(l)$. 
Unlike previous works~\cite{baldrati2022conditioned}, we also preserve the fine-grained token-level features in addition to the global representations, which facilitates to explore richer interactions between modalities. For image token-level features, we use a linear layer projecting them to the same d-dimension as text modality representations $(d=512)$. Note that the features are actually a set of token-level features and projected global representations, they can be formulated in a unified way as follows:
\begin{equation}
V=\{\mathrm{proj}(\boldsymbol v_{cls}), \boldsymbol v_1, \boldsymbol v_2, ..., \boldsymbol v_M\},
\label{feature_defination}
\end{equation}
where $M$ is the sequence length or the number of tokens and tokens in V can be from $r,l,t$ and thus produces $I_r, I_{t}$ or $L$.
\subsection{Cross-Modal Decompose}
\label{sec:degradation}
As mentioned in the introduction section, the text as a description to fuse with the reference image leads to ignoring important clues in the reference image. Our goal is to use text (as an instruction) to first guide the model shifts back to a visual prototype of the image (features shared in the reference image and the target image). 

Inspired by the token selection~\cite{liu2022ts2} in vision transformers, we propose a trainable cross-modal decompose module to direct the semantics within the modality towards the two conceptual directions. For example, given a modification text ``\emph{Dog is barking, nothing blocking his mouth}'', it implies a dog from ``\emph{(not barking, blocked) to (barking)}''. With the help of reference images, we can distinguish this set of opposite semantic information. As shown in Figure~\ref{fig:overview}, the inputs are a set of tokens from the provided text $\{\boldsymbol v_1^l, \boldsymbol v_2^l, ..., \boldsymbol v_M^l\}\in \mathbb{R}^{M\times d}$ ($M$ is the text sequence length), we first concatenate them with the global semantic representation of the reference image $I_{r}^g=\mathrm{proj}(\boldsymbol v_{cls}^r)$. Then we feed the concatenated features $X^l=\{\boldsymbol x_1,\boldsymbol x_2,...,\boldsymbol x_M\}$, $\boldsymbol x_i=[I_{r}^g,\boldsymbol v_i^l]\in \mathbb{R}^{M\times2d}$ ($[\cdot]$ is the concatenation operation) to one MLP Layer followed by a sigmoid activation function. This indicates which conceptual direction each token belongs to:
\begin{equation}
\begin{aligned}
   C_l=\mathrm{sigmoid}(W_lX^l+b_l)\in \mathbb{R}^{M\times 1}
\end{aligned}
\label{Eq:decomposing text}
\end{equation}
Multiplied with the original token-level text features, we generate the positive and negative guiding text representations.
\begin{equation}
\begin{aligned}
    L^+&=C_l\odot L\\
    L^-&=(1-C_l)\odot L,\\
\end{aligned}
\label{Eq:decomposed text}
\end{equation}
So far, we have obtained the explicit guidance signal about the text. Conditioned on this, to determine which among the reference images should be kept or discarded, we proceed similarly but exchange the positions of the two modalities.
\begin{equation}
\begin{aligned}
X^r&=\{\boldsymbol x_1,\boldsymbol x_2,...,\boldsymbol x_P\},\boldsymbol x_i=[L^-,\boldsymbol v_i^r]\in \mathbb{R}^{P\times2d} \\
   C_r&=\mathrm{sigmoid}(W_rX^r+b_r) \in \mathbb{R}^{P\times 1},
\end{aligned}
\label{Eq:decomposing img}
\end{equation}
where $P$ is the number of patches and $X^r$ is the concatenated features of reference image tokens $\{\boldsymbol v_1^r, \boldsymbol v_2^r, ..., \boldsymbol v_P^r\}$ and global semantic representations of $L^-$. After that, we obtain $I_r^0$ by
\begin{equation}
\begin{aligned}
    I_{r}^0&=C_r\odot I_{r}\\
    \overline{I}_{r}^0&=(1-C_r)\odot I_{r},\\
\end{aligned}
\label{I_ref}
\end{equation}
where $I_{r}^0$ represents the expected visual prototypes, which preserves the core information of the given reference image. 
\subsection{Late Composition}
\label{sec:upgradation}
Previous works design a variety of methods to learn joint representations from composed image-text queries, ie, gated residual fusion and attention-based fusion. 
Taking the global representations as residuals can effectively prevent gradient vanishing. 

In our late composition module, there are two parallel branches, one for processing positive guiding text and visual prototypes, and the other for negative guidance and irrelevant features in reference images. In each branch, we first add modality-specific embeddings for inputs from different modalities, similar to the segment embedding in BERT~\cite{DBLP:conf/naacl/BERT}. Then we fuse them via the transformer encode block $\mathcal{F(\cdot)}$. In detail, take the top branch as an example, given token sequences $[L^+,I_{r}^0]$, the fused features represent as:
\begin{equation}
F_{en}=\mathcal{F}([L^++E_l,I_{r}^0+E_i]),\\
\label{attn_fusion}
\end{equation}
where $E_l,E_i$ is the text modality embedding and image modality embedding respectively. The same fusion as in Eq.(\ref{attn_fusion}) is performed in the bottom branch for the inputs $\{L^-,\overline{I}_{r}^0\}$. Note that the features of the target image also go through the fusion layer but without the input of the text modality. 

Finally following the work of~\cite{baldrati2022conditioned}, the final predicted feature $F_p$ is a linear addition between the convex combination of the global reference image $I_{r}^g$ and global text features $L$ and the learned pooled fused features $\hat{F_{en}}$. We denoted the final fused features in the top branch as $F_p^+$ and the fused features in the bottom branch as $F_p^-$.
\begin{table*}[t]
\begin{center}
\begin{tabular}{c|cccc|ccc|c}
\toprule[1.5pt]
\multicolumn{1}{c|}{} &
  \multicolumn{4}{c|}{Recall@K} &
  \multicolumn{3}{c|}{ $\mathrm{Recall_{subset}@K}$} &
  \multicolumn{1}{c}{} \\ \cline{2-8}
\multicolumn{1}{c|}{\multirow{-2}{*}{Method}} &
  \multicolumn{1}{c|}{K=1} &
  \multicolumn{1}{c|}{K=5} &
  \multicolumn{1}{c|}{K=10} &
  \multicolumn{1}{c|}{K=50} &
  \multicolumn{1}{c|}{K=1} &
  \multicolumn{1}{c|}{K=2} &
  \multicolumn{1}{c|}{K=3} &
  \multicolumn{1}{c}{\multirow{-2}{*}{(R@5+R\_sub@1)/2}} \\ \hline\hline
TIRG~\cite{vo2019composing} &
  14.61 &
  48.37 &
  64.08 &
  90.03 &
  22.67 &
  44.97 &
  65.14 &
  35.52 \\
TIRG+LastConv~\cite{vo2019composing} &
  11.04 &
  35.68 &
  51.27 &
  83.29 &
  23.82 &
  45.65 &
  64.55 &
  29.75 \\
MAAF~\cite{dodds2020modality} &
  10.31 &
  33.03 &
  48.30 &
  80.06 &
  21.05 &
  41.81 &
  61.60 &
  27.04 \\
MAAF+BERT~\cite{dodds2020modality} &
  10.12 &
  33.10 &
  48.01 &
  80.57 &
  22.04 &
  42.41 &
  62.14 &
  27.57 \\
MAAF-IT~\cite{dodds2020modality} &
  9.90 &
  32.86 &
  48.83 &
  80.27 &
  21.17 &
  42.04 &
  60.91 &
  27.02 \\
MAAF-RP~\cite{dodds2020modality} &
  10.22 &
  33.32 &
  48.68 &
  81.84 &
  21.41 &
  42.17 &
  61.60 &
  27.37 \\
ARTEMIS~\cite{delmas2022artemis} &
  16.96 &
  46.10 &
  61.31 &
  87.73 &
  39.99 &
  62.20 &
  75.67 &
  43.05 \\
CIRPLANT~\cite{DBLP:conf/iccv/0002OTG21} &
  15.18 &
  43.36 &
  60.48 &
  87.64 &
  33.81 &
  56.99 &
  75.40 &
  38.59 \\
CIRPLANT w/OSCAR~\cite{DBLP:conf/iccv/0002OTG21} &
  19.55 &
  52.55 &
  68.39 &
  92.38 &
  39.20 &
  63.03 &
  79.49 &
  45.88 \\
CLIP4Cir~\cite{baldrati2022conditioned} &
  38.53 &
  69.98 &
  81.86 &
  95.93 &
  68.19 &
  85.64 &
  94.17 &
  69.09 \\
  \hline
SSN &
  \textbf{43.91} &
  \textbf{77.25} &
  \textbf{86.48} &
  \textbf{97.45} &
  \textbf{71.76} &
  \textbf{88.63} &
  \textbf{95.54} &
  \textbf{74.51} \\
\bottomrule[1.5pt]
\end{tabular}
\end{center}
\caption{Comparisons with the state-of-the-art methods for composed image retrieval on the CIRR dataset. Here we show all Recall@K, $\mathrm{Recall_{subset}@K}$ and the mean recall. Our complete SSN model obtain significant improvement compared to other sota methods. The best results are in bold.}
\label{table1:cirr_sota}
\end{table*}
\begin{table*}[]
\begin{center}
\begin{tabular}{c|cc|cc|cc|ccc}
\toprule[1.5pt]
\multicolumn{1}{c|}{\multirow{2}{*}{Method}} &
  \multicolumn{2}{c|}{Tops\&Tees} &
  \multicolumn{2}{c|}{Dress} &
  \multicolumn{2}{c|}{Shirt} &
  \multicolumn{3}{c}{Average} \\ \cline{2-10} 
\multicolumn{1}{c|}{} &
  \multicolumn{1}{l|}{R@10} &
  \multicolumn{1}{l|}{R@50} &
  \multicolumn{1}{l|}{R@10} &
  \multicolumn{1}{l|}{R@50} &
  \multicolumn{1}{l|}{R@10} &
  \multicolumn{1}{l|}{R@50} &
  \multicolumn{1}{l|}{R@10} &
  \multicolumn{1}{l|}{R@50} &
  \multicolumn{1}{c}{mean} \\ \hline\hline
TIRG~\cite{vo2019composing}& 19.08 & 39.62 & 14.87 & 34.66 & 18.26 & 37.89 & 17.40 & 37.39 & 27.40 \\
JVSM~\cite{Chen2020LearningJV}& 13.00 & 26.90 & 10.70 & 25.90 & 12.00 & 27.10 & 11.90 & 26.60 & 19.25 \\
VAL~\cite{chen2020image}& 27.53 & 51.68 & 22.53 & 44.00 & 22.38 & 44.15 & 24.15 & 46.61 & 35.38 \\
CoSMo~\cite{lee2021cosmo}& 29.21 & 57.46 & 25.64 & 50.30 & 24.90 & 49.18 & 26.58 & 53.21 & 39.90 \\
CLVC-Net~\cite{wen2021comprehensive}& 33.50 & 64.00 & 29.85 & 56.47 & 28.75 & 54.76 & 30.70 & 58.41 & 44.56 \\
SAC~\cite{DBLP:conf/wacv/JandialBCCSK22}& 32.70 & 61.23 & 26.52 & 51.01 & 28.02 & 51.86 & 29.08 & 54.70 & 41.89 \\
DCNet~\cite{kim2021dual}& 30.44 & 58.29 & 28.95 & 56.07 & 23.95 & 47.30 & 27.78 & 53.89 & 40.84 \\
MAAF~\cite{dodds2020modality}& 27.90 & 53.60 & 23.80 & 48.60 & 21.30 & 44.20 & 24.30 & 48.80 & 36.55 \\
CIRPLANT~\cite{DBLP:conf/iccv/0002OTG21}& 21.64 & 45.38 & 17.45 & 40.41 & 17.53 & 38.81 & 18.87 & 41.53 & 30.20 \\
ARTEMIS~\cite{delmas2022artemis}& 29.20 & 54.83 & 27.16 & 52.40 & 21.78 & 43.64 & 26.05 & 50.29 & 38.17 \\
MUR~\cite{chen2022composed}& 37.37 & 68.41 & 30.60 & 57.46 & 31.54 & 58.29 & 33.17 & 61.39 & 47.28 \\
CLIP4Cir~\cite{baldrati2022conditioned}& 41.41 & 65.37 & 33.81 & 59.40 & \textbf{39.99} & 60.45 & 38.32 & 61.74 & 50.03 \\
\hline
SSN&
  \textbf{44.26} &
  \textbf{69.05} &
  \textbf{34.36} &
  \textbf{60.78} &
  38.13 &
  \textbf{61.83} &
  \textbf{38.92} &
  \textbf{63.89} &
  \textbf{51.40}     \\
\bottomrule[1.5pt]
\end{tabular}
\end{center}
\caption{Comparisons with the state-of-the-art methods for composed image retrieval on the FashionIQ dataset. Here we show all Recall@10 and Recall@50 across all categories. Our complete SSN model outperforms other state-of-the-art methods on most of the metrics. The best result is in bold.}
\label{table1:fashioniq_sota}
\end{table*}
\subsection{Training and Inference}
\label{sec:trainandinfer}
As shown in Figure~\ref{fig:overview}, we use the inner product to measure the similarity between the predicted features and the target image representation and then obtain the ranking list of candidate images. Following~\cite{baldrati2022conditioned,zhao2022progressive,wang2022exploring}, the objective is to minimize batched-based classification loss as follows:
\begin{equation}
\mathcal{L}_c=\frac{1}{B}\sum_{i=1}^B-\mathrm{log}\frac{\mathrm{exp}(\lambda*s(F_p^{(i)},F_{tg}^{(i)}))}{\sum_{j=1}^B\mathrm{exp}(\lambda*s(F_p^{(i)},F_{tg}^{(j)}))},
\label{Eq:retrieval_loss}
\end{equation}
where $\lambda$ is a temperature parameter. Given the two sets of predicted features output by the late composite module, we obtain their similarity distribution to the target image. $\mathrm{z}^-=\mathrm{softmax}(sim(F_p^-,F_{tg})),\mathrm{z}^+=\mathrm{softmax}(sim(F_p^+,F_{tg}))$. Finally, we employ a Kullback-Leibler Divergence loss as a regularization constraint:
\begin{equation}
\mathcal{L}_k=\mathrm{KL}(\mathrm{z}^+\|\mathrm{z}^{gt})-\mathrm{KL}(\mathrm{z}^-\|\mathrm{z}^+)\\
\label{Eq:kl_loss}
\end{equation}
We aim to push away the distance between $\mathrm{z}^+$ and $\mathrm{z}^-$ and thus optimize decompose learning. The overall training loss is described as follows:
\begin{equation}
\mathcal{L}=\mathcal{L}_c+\mathcal{L}_k,
\label{Eq:total_loss}
\end{equation}
It is worth noting that the bottom branch in Figure~\ref{fig:overview} used for fusing $L^-$ and $\overline{I}_{r}^0$ is only for training. During inference, we only use $F_p^+$, the composite features of semantic shifts and visual prototypes to retrieve the target image.
\begin{table*}[]
\begin{center}
\begin{tabular}{c|cccc|ccc|c}
\toprule[1.5pt]
\multicolumn{1}{c|}{} &
  \multicolumn{4}{c|}{Recall@K} &
  \multicolumn{3}{c|}{Recall\_subset@K} &
  \multicolumn{1}{c}{} \\ \cline{2-8}
\multicolumn{1}{c|}{\multirow{-2}{*}{Method}} &
  K=1 &
  K=5 &
  K=10 &
  K=50 &
  K=1 &
  K=2 &
  \multicolumn{1}{c|}{K=3} &
  \multicolumn{1}{c}{\multirow{-2}{*}{(R@5+R\_sub@1)/2}} \\ \cline{1-1} \cline{9-9} 
  \hline\hline
Baseline &
  42.62 &
  76.7 &
  87.06 &
  97.54 &
  68.98 &
  86.73 &
  94.16 &
  72.84 \\
SSN($\mathrm{I_r,L}$) &
  43.34 &
  76.97 &
  87.4 &
  97.2 &
  72.18 &
  88.42 &
  95.24 &
  74.575 \\
  \hline
SSN($\mathrm{I_r^0,L}$) &
  44.43 &
  77.44 &
  86.92 &
  96.96 &
  71.63 &
  88.21 &
  94.88 &
  74.535 \\
SSN($\mathrm{I_r,L^+}$) &
  43.46 &
  77.64 &
  87.51 &
  97.42 &
  71.9 &
  87.9 &
  95.12 &
  74.77 \\
  \hline
\textbf{SSN} &
  \textbf{45.13} &
  \textbf{77.49} &
  \textbf{87.75} &
  \textbf{97.32} &
  \textbf{73.04} &
  \textbf{88.64} &
  \textbf{95.17} &
  \textbf{75.265}\\
  \bottomrule[1.5pt]
\end{tabular}
\end{center}
\caption{Ablation Studies of our SSN model with different components and various settings for decomposition outputs. We report all Recall@K, $\mathrm{Recall_{subset}@K}$, and the mean recall on the validation set of the CIRR dataset.}
\label{table1:decompose_ablation}
\end{table*}
\section{Experiments}
\subsection{Datasets and Metrics}
To verify the effectiveness of our model, we conduct experiments on two standard CIR datasets, involving various domains.

\noindent\textbf{CIRR Dataset} (Compose Image Retrieval on Real-life images)~\cite{DBLP:conf/iccv/0002OTG21} is the first released dataset of open-domain for the CIR task. Each triplet consists of real-life images with human-generated modification sentences. The real-life images come from the popular $\mathrm{NLVR}^2$ dataset~\cite{suhr2018corpus}, which contains real-world entities with reasonable complexity. In 36,554 triplets, 80\% are for training, 10\% are for validation, and 10\% are for evaluation. 

Besides, the authors in~\cite{DBLP:conf/iccv/0002OTG21} also construct subsets of images, which are composed of six negative examples with high visual similarity. This helps to evaluate the performance of the model in true negative images.

\noindent\textbf{FashionIQ Dataset}~\cite{wu2021fashion} is a realistic dataset for interactive image retrieval in the fashion domain. Each query is composed of one reference image and two natural language descriptions about the visual differences of the target image. Following~\cite{baldrati2022conditioned,kim2021dual}, we use the original evaluation split, which includes 5,373, 3,817, and 6,346 images for three specific fashion categories: Tops\&Tees, Dresses, Shirts.

\noindent\textbf{Metrics}. Following previous works~\cite{baldrati2022conditioned, delmas2022artemis,zhao2022progressive}, we employ Recall within top-K as the retrieval performance, which indicates the ratio of the ground-truth target image in the top-K ranking list that is correctly retrieved. For the CIRR dataset, Recall@1, 5, 10, 50 is reported. When only considering images in the subset of the query image, $\mathrm{Recall_{subset}}$@1, 2, 3 is also reported. Recall@5 and $\mathrm{Recall_{subset}@1}$, their mean values, as the mean metric, represent the ability to include the correct image within a smaller ranking list. For the FashionIQ dataset, the value of K is 10 and 50. The average of Recall@10 and Recall@50 is finally computed as a mean metric to evaluate overall retrieval performance.
\subsection{Implementation Details}
We utilize the CLIP~\cite{radford2021learning} model to initialize the image encoder with ViT-B/32. The hidden dimension of the 1-layer 8-head transformer encoder is set to 512. The temperature $\lambda$ of the main retrieval loss (in Eq.(\ref{Eq:retrieval_loss})) is equal to 100. 
Note that for FashionIQ, we fix the image encoder after one training epoch and fine-tune the text encoder only. We adopt AdamW optimizer with an initial learning rate of 5e-5 to train the whole model.  We apply the step scheduler to decay the learning rate by 10 every 10 epochs. The batch size is set to 128 and the network is trained for 50 epochs. All experiments can be implemented with PyTorch on a single NVIDIA RTX 3090 Ti GPU.
\subsection{Comparison with State-of-the-Arts}
\noindent\textbf{Results on CIRR dataset} are presented in Table~\ref{table1:cirr_sota} for the test set. Our model which learns to decompose visual prototypes and semantic shift outperforms the state-of-the-art in all metrics for different backbones. Compared to CLIP4Cir~\cite{baldrati2022conditioned}, a strong competitor that has recently successfully applied the CLIP model to the CIR task, our model outperforms it by 5.42\% mean recall $\mathrm{(R@5+R_{sub}@1})/2$ and increases up to 5.38\% in Top-1 recall metrics. As shown in Table~\ref{table1:cirr_sota}, we also outperform other methods by a large margin. 

\noindent\textbf{Results on FashionIQ dataset} are reported in Table~\ref{table1:fashioniq_sota} for the validation set. Although the model is not improved as much as on the CIRR dataset, our proposed method achieves state-of-the-art results for all categories. Compared to the strongest method, we improve the mean recall by 1.37\%. The limited improvement is due to the domain gap between the fashion data and the open domain CLIP, the size of the data being small, and the data lack of complexity for fine-tuning the CLIP image encoder.
\begin{figure*}
\begin{center}
\includegraphics[width=\linewidth]{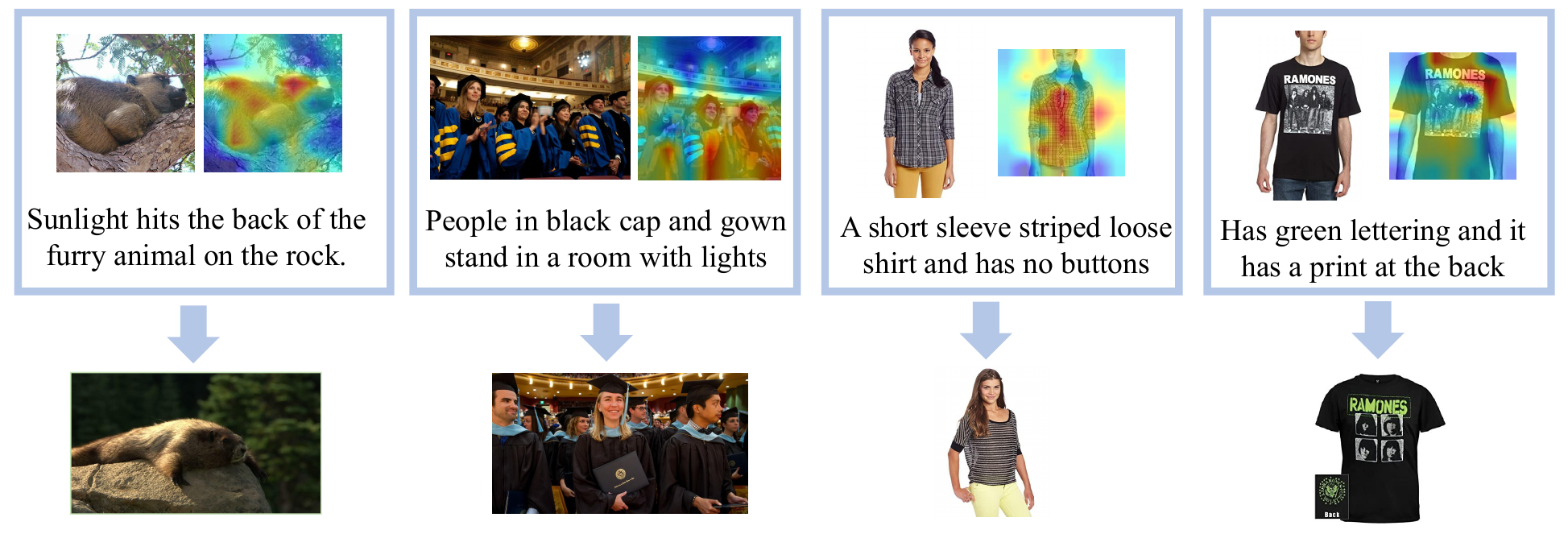}
\end{center}
   \caption{Visualization of where to be concerned when picturing visual prototypes from reference images, in the form of heatmaps. The left two groups are from the CIRR dataset and the right two groups are from the FashionIQ dataset. In the reference image in the first column, the back of the furry animal is highlighted in the visual prototype, indicating the main characteristics of the target image. In the reference image of the third column, the reference image changes the print (lettering and color) in the front of the T-shirt to bring it back to the visual prototype. This suggests that we are not just concerned with the salient objects in the image and that the visual prototype contains a rich set of visual cues.}
\label{fig:cirr_heatmaps}
\end{figure*}
\begin{table}[]
\begin{center}
\begin{tabular}{c|c|c|cc|c|c}
\hline
\multicolumn{1}{c|}{\multirow{2}{*}{}} &
  \multicolumn{1}{c|}{\multirow{2}{*}{shared}} &
  \multicolumn{1}{c|}{\multirow{2}{*}{$\mathcal{L}_k$}} &
  \multicolumn{2}{c|}{R@K} &
  \multicolumn{1}{c|}{\multirow{2}{*}{R\_sub@1}} &
  \multicolumn{1}{c}{\multirow{2}{*}{mean}} \\ \cline{4-5}
\multicolumn{1}{c}{} &
  \multicolumn{1}{|c|}{} &
  \multicolumn{1}{c|}{} &
  \multicolumn{1}{c|}{K=1} &
  \multicolumn{1}{c|}{K=5} &
  \multicolumn{1}{c|}{} &
  \multicolumn{1}{c}{} \\ \hline
1 &
  \checkmark &
  \checkmark &
  43.77 &
  77.30 &
  71.66 &
  74.91 \\
2 &
  $\times$ &
  $\times$ &
  44.32 &
  77.42 &
  71.92 &
  74.67 \\
3 &
  $\times$ &
  \checkmark &
  \textbf{45.13} &
  \textbf{77.49} &
  \textbf{73.04} &
  \textbf{75.27}\\
\hline
\end{tabular}
\end{center}
\caption{Ablation experiments on loss function terms $\mathcal{L}_k$ and an exploration of whether two decomposing layers (one for images and one for text) share parameters. We report all metrics on the validation set of CIRR dataset. The model with a regularized loss $\mathcal{L}_k$ (the second row) performs better slightly than the one without (the third row).}
\label{table:kl_shared}
\end{table}
\subsection{Ablation Studies}
\textbf{Model architecture}.
In order to demonstrate the contributions of individual components in our design, we first conducted experiments about ablated models. Moreover, we also explored several various settings for the decomposition outputs that are used during upgradation. Table~\ref{table1:decompose_ablation} presents the detailed results on the validation set of the CIRR dataset. We report all the Recall metrics. The different ablated models are as follows:
\begin{itemize}
\item \textbf{Baseline}: it is the model without any designed module.
\item \textbf{SSN($\mathrm{I_r,L}$)}: it is the SSN model without degradation. That means the inputs for the upgradation are original dense tokens extracted from the visual and textual encoder.
\item \textbf{SSN($\mathrm{I_r^0,L}$)}: it is the complete SSN model where visual prototypes and original text features are decomposition outputs. 
\item \textbf{SSN($\mathrm{I_r,L^+}$)}: it is the complete SSN model where the original reference image and positive guiding text features are decomposition outputs. 
\item \textbf{SSN}: it is the full SSN model where positive guiding text features ($\mathrm{L^+}$) rich the degraded visual prototypes ($\mathrm{I_r^0}$) during upgradation and then produces the final representations. 
\end{itemize}
There are three following observations in Table~\ref{table1:decompose_ablation}:  1) the SSN($\mathrm{I_r,L}$) model slightly outperforms the baseline by 0.72\% in Recall@1 because of fine-grained token features. Our proposed method (SSN) achieves the best performance and gains a more significant improvement over the baseline model (2.51\% in Recall@1). This highlights the effectiveness of decomposing semantic shifts into two steps. 2) SSN($\mathrm{I_r,L^+}$) is comparable to SSN ($\mathrm{I_r,L}$) model. This is because when only decomposing text instructions without generating visual prototypes, semantic shifts lack a well-acted object. 3) Based on the SSN ($\mathrm{I_r,L}$) model, two models (SSN($\mathrm{I_r^0,L}$) \& SSN) picturing reference image to visual prototypes achieved further improvements up to 1.79\%. This supports the motivation discussed in Section~\ref{sec:intro} that the visual cue of the reference images should not be disregarded. 

\noindent\textbf{Loss fubction}.
Our total training objective in Eq.(\ref{Eq:total_loss}) involves two aspects: the main retrieval loss and additional regularized loss $\mathcal{L}_k$. To demonstrate the effectiveness of the regularized constraint decomposing features, we performed ablation experiments on $\mathcal{L}_k$ in Eq.(\ref{Eq:total_loss}). Comparing the second and third rows in Table~\ref{table:kl_shared}, we observe that the model with a regularized loss $\mathcal{L}_k$ performs better than the one without, despite only a slight improvement. This shows additional loss $\mathcal{L}_k$ helps to learn optimal decomposed features from original CLIP representations.

\noindent\textbf{Shared Parameter of decomposing layer for image and text?}
In Figure~\ref{fig:overview}, we employ the same structure: one MLP layer followed by a sigmoid activation function, to decompose the semantic shifts and obtain visual prototypes. To explore whether components with the same structure can share parameters, we conduct additional experiments. From rows one and three in Table~\ref{table:kl_shared}, we can see that MLP layers that share parameters hurt the performance of the model. This is because decomposing networks for reference images and text have different objectives despite the same architecture. Producing visual prototypes is to get invariant features (shared with target images) while decomposing semantics in the text instruction is to get semantically shifted features.
\subsection{Qualitative Results}
As shown in Figure~\ref{fig:cirr_heatmaps}, we visualize what details are retained in the process of picturing visual prototypes from reference images on both datasets. In Section~\ref{sec:degradation}, we generate $C_r$ to indicate whether certain features of the reference image are preserved or not. The heatmap is a merging of $C_r$ and the original reference images. From the heatmaps, we can tell where the visual prototype and the original reference image have changed and the extent of these changes. With decomposing semantic shifts as guidance, we observe that the majority of important information in the reference image receives more attention. 
\section{Conclusion}
In this paper, we focus on the composed image retrieval task, an extended image retrieval task. Given the provided reference image and text requirements pair, the goal is to retrieve the desired target image. We first rethink the text as an instruction and then propose a Semantic Shift Network (SSN) to decompose the text instructions into degradation and upgradation. The text first directs the reference image toward the visual prototype and then guides the visual prototype closer to the target image. Extensive experiments on two benchmark datasets verify the effectiveness of the proposed method and show that our model significantly outperforms state-of-the-art methods by 5.42\% and 1.37\% on the mean of Recall@K, respectively. In the future, we intend to explore other complex mechanisms to model the text instruction in the CIR task and extend to the zero-shot setting.


\begin{thebibliography}{35}
\providecommand{\natexlab}[1]{#1}

\bibitem[{Baldrati et~al.(2022)Baldrati, Bertini, Uricchio, and
  Del~Bimbo}]{baldrati2022conditioned}
Baldrati, A.; Bertini, M.; Uricchio, T.; and Del~Bimbo, A. 2022.
\newblock Conditioned and composed image retrieval combining and partially
  fine-tuning CLIP-based features.
\newblock In \emph{Proceedings of the IEEE/CVF Conference on Computer Vision
  and Pattern Recognition}, 4959--4968.

\bibitem[{Chen et~al.(2022{\natexlab{a}})Chen, Liu, Wang, Bakker, Georgiou,
  Fieguth, Liu, and Lew}]{chen2022deep}
Chen, W.; Liu, Y.; Wang, W.; Bakker, E.~M.; Georgiou, T.; Fieguth, P.; Liu, L.;
  and Lew, M.~S. 2022{\natexlab{a}}.
\newblock Deep learning for instance retrieval: A survey.
\newblock \emph{IEEE Transactions on Pattern Analysis and Machine
  Intelligence}.

\bibitem[{Chen and Bazzani(2020)}]{Chen2020LearningJV}
Chen, Y.; and Bazzani, L. 2020.
\newblock Learning Joint Visual Semantic Matching Embeddings for
  Language-Guided Retrieval.
\newblock In \emph{European Conference on Computer Vision}.

\bibitem[{Chen, Gong, and Bazzani(2020)}]{chen2020image}
Chen, Y.; Gong, S.; and Bazzani, L. 2020.
\newblock Image search with text feedback by visiolinguistic attention
  learning.
\newblock In \emph{Proceedings of the IEEE/CVF Conference on Computer Vision
  and Pattern Recognition}, 3001--3011.

\bibitem[{Chen et~al.(2022{\natexlab{b}})Chen, Zheng, Ji, Qu, and
  Chua}]{chen2022composed}
Chen, Y.; Zheng, Z.; Ji, W.; Qu, L.; and Chua, T.-S. 2022{\natexlab{b}}.
\newblock Composed Image Retrieval with Text Feedback via Multi-grained
  Uncertainty Regularization.
\newblock \emph{arXiv preprint arXiv:2211.07394}.

\bibitem[{Delmas et~al.(2022)Delmas, de~Rezende, Csurka, and
  Larlus}]{delmas2022artemis}
Delmas, G.; de~Rezende, R.~S.; Csurka, G.; and Larlus, D. 2022.
\newblock Artemis: Attention-based retrieval with text-explicit matching and
  implicit similarity.
\newblock \emph{arXiv preprint arXiv:2203.08101}.

\bibitem[{Deng et~al.(2018)Deng, Wu, Wu, Hu, Lyu, and Tan}]{deng2018visual}
Deng, C.; Wu, Q.; Wu, Q.; Hu, F.; Lyu, F.; and Tan, M. 2018.
\newblock Visual grounding via accumulated attention.
\newblock In \emph{Proceedings of the IEEE conference on computer vision and
  pattern recognition}, 7746--7755.

\bibitem[{Devlin et~al.(2019)Devlin, Chang, Lee, and
  Toutanova}]{DBLP:conf/naacl/BERT}
Devlin, J.; Chang, M.; Lee, K.; and Toutanova, K. 2019.
\newblock {BERT:} Pre-training of Deep Bidirectional Transformers for Language
  Understanding.
\newblock In Burstein, J.; Doran, C.; and Solorio, T., eds., \emph{Proceedings
  of the 2019 Conference of the North American Chapter of the Association for
  Computational Linguistics: Human Language Technologies, {NAACL-HLT} 2019,
  Minneapolis, MN, USA, June 2-7, 2019, Volume 1 (Long and Short Papers)},
  4171--4186. Association for Computational Linguistics.

\bibitem[{Devlin et~al.(2018)Devlin, Chang, Lee, and
  Toutanova}]{devlin2018bert}
Devlin, J.; Chang, M.-W.; Lee, K.; and Toutanova, K. 2018.
\newblock Bert: Pre-training of deep bidirectional transformers for language
  understanding.
\newblock \emph{arXiv preprint arXiv:1810.04805}.

\bibitem[{Dodds et~al.(2020)Dodds, Culpepper, Herdade, Zhang, and
  Boakye}]{dodds2020modality}
Dodds, E.; Culpepper, J.; Herdade, S.; Zhang, Y.; and Boakye, K. 2020.
\newblock Modality-agnostic attention fusion for visual search with text
  feedback.
\newblock \emph{arXiv preprint arXiv:2007.00145}.

\bibitem[{Goenka et~al.(2022{\natexlab{a}})Goenka, Zheng, Jaiswal, Chada, Wu,
  Hedau, and Natarajan}]{DBLP:conf/cvpr/GoenkaZJC0HN22}
Goenka, S.; Zheng, Z.; Jaiswal, A.; Chada, R.; Wu, Y.; Hedau, V.; and
  Natarajan, P. 2022{\natexlab{a}}.
\newblock FashionVLP: Vision Language Transformer for Fashion Retrieval with
  Feedback.
\newblock In \emph{{IEEE/CVF} Conference on Computer Vision and Pattern
  Recognition, {CVPR} 2022}, 14085--14095. {IEEE}.

\bibitem[{Goenka et~al.(2022{\natexlab{b}})Goenka, Zheng, Jaiswal, CHADA, Wu,
  Hedau, and Natarajan}]{Goenka2022}
Goenka, S.; Zheng, Z.; Jaiswal, A.; CHADA, R.; Wu, Y.; Hedau, V.; and
  Natarajan, P. 2022{\natexlab{b}}.
\newblock FashionVLP: Vision language transformer for fashion retrieval with
  feedback.
\newblock In \emph{CVPR 2022}.

\bibitem[{Goenka et~al.(2022{\natexlab{c}})Goenka, Zheng, Jaiswal, Chada, Wu,
  Hedau, and Natarajan}]{DBLP:conf/cvpr/fashionvlp}
Goenka, S.; Zheng, Z.; Jaiswal, A.; Chada, R.; Wu, Y.; Hedau, V.; and
  Natarajan, P. 2022{\natexlab{c}}.
\newblock FashionVLP: Vision Language Transformer for Fashion Retrieval with
  Feedback.
\newblock In \emph{{IEEE/CVF} Conference on Computer Vision and Pattern
  Recognition, {CVPR} 2022}, 14085--14095. {IEEE}.

\bibitem[{Hosseinzadeh and Wang(2020{\natexlab{a}})}]{hosseinzadeh2020composed}
Hosseinzadeh, M.; and Wang, Y. 2020{\natexlab{a}}.
\newblock Composed query image retrieval using locally bounded features.
\newblock In \emph{Proceedings of the IEEE/CVF Conference on Computer Vision
  and Pattern Recognition}, 3596--3605.

\bibitem[{Hosseinzadeh and
  Wang(2020{\natexlab{b}})}]{DBLP:conf/cvpr/HosseinzadehW20}
Hosseinzadeh, M.; and Wang, Y. 2020{\natexlab{b}}.
\newblock Composed Query Image Retrieval Using Locally Bounded Features.
\newblock In \emph{2020 {IEEE/CVF} Conference on Computer Vision and Pattern
  Recognition, {CVPR} 2020, Seattle, WA, USA, June 13-19, 2020}, 3593--3602.
  Computer Vision Foundation / {IEEE}.

\bibitem[{Jandial et~al.(2022)Jandial, Badjatiya, Chawla, Chopra, Sarkar, and
  Krishnamurthy}]{DBLP:conf/wacv/JandialBCCSK22}
Jandial, S.; Badjatiya, P.; Chawla, P.; Chopra, A.; Sarkar, M.; and
  Krishnamurthy, B. 2022.
\newblock {SAC:} Semantic Attention Composition for Text-Conditioned Image
  Retrieval.
\newblock In \emph{{IEEE/CVF} Winter Conference on Applications of Computer
  Vision, {WACV} 2022}, 597--606. {IEEE}.

\bibitem[{Kim et~al.(2021)Kim, Yu, Kim, and Kim}]{kim2021dual}
Kim, J.; Yu, Y.; Kim, H.; and Kim, G. 2021.
\newblock Dual compositional learning in interactive image retrieval.
\newblock In \emph{Proceedings of the AAAI Conference on Artificial
  Intelligence}, volume~35, 1771--1779.

\bibitem[{Kim, Son, and Kim(2021)}]{kim2021vilt}
Kim, W.; Son, B.; and Kim, I. 2021.
\newblock Vilt: Vision-and-language transformer without convolution or region
  supervision.
\newblock In \emph{International Conference on Machine Learning}, 5583--5594.
  PMLR.

\bibitem[{Lee, Kim, and Han(2021)}]{lee2021cosmo}
Lee, S.; Kim, D.; and Han, B. 2021.
\newblock Cosmo: Content-style modulation for image retrieval with text
  feedback.
\newblock In \emph{Proceedings of the IEEE/CVF Conference on Computer Vision
  and Pattern Recognition}, 802--812.

\bibitem[{Liu et~al.(2022{\natexlab{a}})Liu, Wang, Yang, Zhou, Yao, Shao, and
  Zhao}]{liu2022show}
Liu, B.; Wang, D.; Yang, X.; Zhou, Y.; Yao, R.; Shao, Z.; and Zhao, J.
  2022{\natexlab{a}}.
\newblock Show, deconfound and tell: Image captioning with causal inference.
\newblock In \emph{Proceedings of the IEEE/CVF Conference on Computer Vision
  and Pattern Recognition}, 18041--18050.

\bibitem[{Liu et~al.(2022{\natexlab{b}})Liu, Xiong, Xu, Cao, and
  Jin}]{liu2022ts2}
Liu, Y.; Xiong, P.; Xu, L.; Cao, S.; and Jin, Q. 2022{\natexlab{b}}.
\newblock Ts2-net: Token shift and selection transformer for text-video
  retrieval.
\newblock In \emph{Computer Vision--ECCV 2022: 17th European Conference, Tel
  Aviv, Israel, October 23--27, 2022, Proceedings, Part XIV}, 319--335.
  Springer.

\bibitem[{Liu et~al.(2021)Liu, Opazo, Teney, and
  Gould}]{DBLP:conf/iccv/0002OTG21}
Liu, Z.; Opazo, C.~R.; Teney, D.; and Gould, S. 2021.
\newblock Image Retrieval on Real-life Images with Pre-trained
  Vision-and-Language Models.
\newblock In \emph{2021 {IEEE/CVF} International Conference on Computer Vision,
  {ICCV}}, 2105--2114. {IEEE}.

\bibitem[{Lu et~al.(2019)Lu, Batra, Parikh, and Lee}]{lu2019vilbert}
Lu, J.; Batra, D.; Parikh, D.; and Lee, S. 2019.
\newblock Vilbert: Pretraining task-agnostic visiolinguistic representations
  for vision-and-language tasks.
\newblock \emph{Advances in neural information processing systems}, 32.

\bibitem[{Radford et~al.(2021)Radford, Kim, Hallacy, Ramesh, Goh, Agarwal,
  Sastry, Askell, Mishkin, Clark et~al.}]{radford2021learning}
Radford, A.; Kim, J.~W.; Hallacy, C.; Ramesh, A.; Goh, G.; Agarwal, S.; Sastry,
  G.; Askell, A.; Mishkin, P.; Clark, J.; et~al. 2021.
\newblock Learning transferable visual models from natural language
  supervision.
\newblock In \emph{International conference on machine learning}, 8748--8763.
  PMLR.

\bibitem[{Suhr et~al.(2018)Suhr, Zhou, Zhang, Zhang, Bai, and
  Artzi}]{suhr2018corpus}
Suhr, A.; Zhou, S.; Zhang, A.; Zhang, I.; Bai, H.; and Artzi, Y. 2018.
\newblock A corpus for reasoning about natural language grounded in
  photographs.
\newblock \emph{arXiv preprint arXiv:1811.00491}.

\bibitem[{Sun et~al.(2021)Sun, Shen, Wang, Bao, and Zhou}]{sun2021loftr}
Sun, J.; Shen, Z.; Wang, Y.; Bao, H.; and Zhou, X. 2021.
\newblock LoFTR: Detector-free local feature matching with transformers.
\newblock In \emph{Proceedings of the IEEE/CVF conference on computer vision
  and pattern recognition}, 8922--8931.

\bibitem[{Vo et~al.(2019)Vo, Jiang, Sun, Murphy, Li, Fei-Fei, and
  Hays}]{vo2019composing}
Vo, N.; Jiang, L.; Sun, C.; Murphy, K.; Li, L.-J.; Fei-Fei, L.; and Hays, J.
  2019.
\newblock Composing text and image for image retrieval-an empirical odyssey.
\newblock In \emph{Proceedings of the IEEE/CVF conference on computer vision
  and pattern recognition}, 6439--6448.

\bibitem[{Wang et~al.(2022)Wang, Nezhadarya, Sadhu, and
  Zhang}]{wang2022exploring}
Wang, C.; Nezhadarya, E.; Sadhu, T.; and Zhang, S. 2022.
\newblock Exploring Compositional Image Retrieval with Hybrid Compositional
  Learning and Heuristic Negative Mining.
\newblock In \emph{Findings of the Association for Computational Linguistics:
  EMNLP 2022}, 1273--1285.

\bibitem[{Wang et~al.(2019)Wang, Liu, Li, Sheng, Yan, Wang, and
  Shao}]{wang2019camp}
Wang, Z.; Liu, X.; Li, H.; Sheng, L.; Yan, J.; Wang, X.; and Shao, J. 2019.
\newblock Camp: Cross-modal adaptive message passing for text-image retrieval.
\newblock In \emph{Proceedings of the IEEE/CVF international conference on
  computer vision}, 5764--5773.

\bibitem[{Weinzaepfel et~al.(2022)Weinzaepfel, Lucas, Larlus, and
  Kalantidis}]{weinzaepfel2022learning}
Weinzaepfel, P.; Lucas, T.; Larlus, D.; and Kalantidis, Y. 2022.
\newblock Learning super-features for image retrieval.
\newblock \emph{arXiv preprint arXiv:2201.13182}.

\bibitem[{Wen et~al.(2021)Wen, Song, Yang, Zhan, and
  Nie}]{wen2021comprehensive}
Wen, H.; Song, X.; Yang, X.; Zhan, Y.; and Nie, L. 2021.
\newblock Comprehensive linguistic-visual composition network for image
  retrieval.
\newblock In \emph{Proceedings of the 44th International ACM SIGIR Conference
  on Research and Development in Information Retrieval}, 1369--1378.

\bibitem[{Wu et~al.(2021)Wu, Gao, Guo, Al-Halah, Rennie, Grauman, and
  Feris}]{wu2021fashion}
Wu, H.; Gao, Y.; Guo, X.; Al-Halah, Z.; Rennie, S.; Grauman, K.; and Feris, R.
  2021.
\newblock Fashion iq: A new dataset towards retrieving images by natural
  language feedback.
\newblock In \emph{Proceedings of the IEEE/CVF Conference on Computer Vision
  and Pattern Recognition}, 11307--11317.

\bibitem[{Yang et~al.(2021)Yang, Wang, Zhou, and Li}]{DBLP:conf/mm/Yang0ZL21}
Yang, Y.; Wang, M.; Zhou, W.; and Li, H. 2021.
\newblock Cross-modal Joint Prediction and Alignment for Composed Query Image
  Retrieval.
\newblock In Shen, H.~T.; Zhuang, Y.; Smith, J.~R.; Yang, Y.; C{\'{e}}sar, P.;
  Metze, F.; and Prabhakaran, B., eds., \emph{{MM} '21: {ACM} Multimedia
  Conference}, 3303--3311. {ACM}.

\bibitem[{Zhang et~al.(2022)Zhang, Yan, Zhang, and
  Xu}]{DBLP:conf/mm/ZhangYZX22}
Zhang, F.; Yan, M.; Zhang, J.; and Xu, C. 2022.
\newblock Comprehensive Relationship Reasoning for Composed Query Based Image
  Retrieval.
\newblock In Magalh{\~{a}}es, J.; Bimbo, A.~D.; Satoh, S.; Sebe, N.;
  Alameda{-}Pineda, X.; Jin, Q.; Oria, V.; and Toni, L., eds., \emph{{MM} '22:
  The 30th {ACM} International Conference on Multimedia}, 4655--4664. {ACM}.

\bibitem[{Zhao, Song, and Jin(2022)}]{zhao2022progressive}
Zhao, Y.; Song, Y.; and Jin, Q. 2022.
\newblock Progressive Learning for Image Retrieval with Hybrid-Modality
  Queries.
\newblock In \emph{Proceedings of the 45th International ACM SIGIR Conference
  on Research and Development in Information Retrieval}, 1012--1021.

\end{thebibliography}

\clearpage
\appendix
This is the supplementary material of Decomposing Semantic Shifts for Composed Image Retrieval. It includes a detailed network and algorithmic procedures.(Appendix~\ref{appendix:network_detailes}), additional experiments about the network (Appendix~\ref{appendix:exp}), retrieval examples on both CIRR and FashionIQ datasets, and more qualitative results for decomposing network on both datasets (Appendix~\ref{appendix:qualitative}). Codes will be publicly available.  
\section{Network Details}
\label{appendix:network_detailes}
\setcounter{table}{0}
\renewcommand\thetable{\arabic{table}}
\setcounter{figure}{0}
\renewcommand\thefigure{\arabic{figure}}
\textbf{Network Design}.
Table~\ref{table:network details} summarizes the architecture of the decomposing and upgrading networks. Note that the inputs for the decomposing network are token-level features in the CLIP model~\cite{radford2021learning}, rather than the final pooled embedding. The extracted pooled representations for reference images and text modifiers will be utilized during the linear addition process. 

\noindent\textbf{Algorithm}.
Algorithm~\ref{alg:1} summarizes the main process of our proposed Semantic Shift Network (SSN).
\begin{algorithm}
	\renewcommand{\algorithmicrequire}{\textbf{Input:}}
	\renewcommand{\algorithmicensure}{\textbf{Output:}}
	\caption{Semantic Shifts Network for Composed Image Retrieval}
	\label{alg:1}
	\begin{algorithmic}[1]
            \REQUIRE reference image $r$, text modifier $l$, target image $t$
            \ENSURE ranked list of target images
            \STATE Extract features of image and text respectively by CLIP image and text encoder $\varPhi_I, \varPhi_L$. \\$I_r=\varPhi_I(r), I_t=\varPhi_I(t), L=\varPhi_L(l)$.
            \STATE Concatenate cross-modal features\\$\boldsymbol v_i^l(\boldsymbol v_j^r)$ are token-level representations\\
            $X^l=\{\boldsymbol x_1,\boldsymbol x_2,...,\boldsymbol x_M\}, \boldsymbol x_i=[I_{r}^g, \boldsymbol v_i^l],$ \\$X^r=\{\boldsymbol x_1,\boldsymbol x_2,...,\boldsymbol x_P\}, \boldsymbol x_j=[L^-, \boldsymbol v_j^r]$.
            \STATE Get decomposing weights information $C_l,C_r$ for $X^l,X^r$ by Eq.(2) and Eq.(4)
            \STATE Decompose semantic shifts by Eq.(3), $L\to \{L^+,L^-\}$
            \STATE Picture the reference image to visual prototype $I_{r}^0$ by Eq.(5)
            \STATE Fuse $I_r^0$ and positive semantics $L^+$ base on transformer encoder $\mathcal{F(\cdot)}\to F_{en}$
            \STATE Get linearly learned pooled token features $\hat{F_{en}}$
            \STATE $F_p^+\gets$ linear addition($\hat{F_{en}}, L, I_r^g$) 
            \STATE Similarity scores $S\gets \mathrm{descending sort}(I_t\otimes F_p^+)$
            \STATE \textbf{return} ranked list
	\end{algorithmic}  
\end{algorithm}

\section{More Experiments}
\label{appendix:exp}
\textbf{Sensitivity Analysis.}
In this section, we do several experiments in order to further demonstrate our SSN model instructs reference images to visual prototypes instead of discarding them. We report Recall@1 and Recall@K on a subset and mean recall metrics since they best capture the model capabilities~\cite{DBLP:conf/iccv/0002OTG21}. We first add a scaled noise $\Delta$ from random sampling to the features of the reference image, which conforms to a Gaussian distribution. The results are reported in Table~\ref{table1:sensitivity_analysis}. As expected, our model drops worse in recall after adding noise when compared to the baseline. This indicates that the proposed SSN model is more sensitive to the reference image features. This also verifies that the core information of the reference image is retained as the prototype and plays an essential role in the retrieval.
\begin{table}[]
\begin{center}
\begin{tabular}{cc|c}
\hline
\multicolumn{1}{c|}{Layer} & Module & \multicolumn{1}{l}{} \\ \hline
\multicolumn{1}{c|}{Image projection} & Linear(768,512) & \multirow{5}{*}{decomposing} \\ \cline{1-2}
\multicolumn{1}{c|}{\multirow{2}{*}{Text decomposing}} & Linear(512,1) &  \\ \cline{2-2}
\multicolumn{1}{c|}{} & Sigmoid &  \\ \cline{1-2}
\multicolumn{1}{c|}{\multirow{2}{*}{Image decomposing}} & Linear(512,1) &  \\ \cline{2-2}
\multicolumn{1}{c|}{} & Sigmoid &  \\ \hline
\multicolumn{1}{c|}{Modality embedding} & Embedding(2,512) & \multirow{6}{*}{upgrading} \\ \cline{1-2}
\multicolumn{1}{c|}{Token-level fusion} & {\makecell[c]{8-head 1-layer \\ transformer encoder}} &  \\ \cline{1-2}
\multicolumn{1}{c|}{\multirow{3}{*}{\makecell[c]{Output token- \\ level   features}}} & Linear(1024,512) &  \\ \cline{2-2}
\multicolumn{1}{c|}{} & Relu &  \\ \cline{2-2}
\multicolumn{1}{c|}{} & Linear(512,512) &  \\ \cline{1-2}
\multicolumn{2}{c|}{Addition} &  \\  \hline
\end{tabular}
\end{center}
\caption{The detailed description of the decomposing and upgrading network. The inputs for the decomposing network are features in the output of the CLIP image and text encoder. The outputs for the upgrading network are optimal composited features. }
\label{table:network details}
\end{table}
\begin{table}
\begin{center}
 \resizebox{\linewidth}{!}{
\begin{tabular}{c|c|c|ccc|c}
\toprule[1.5pt]
\multicolumn{1}{c|}{} &
  \multicolumn{1}{c|}{} &
  \multicolumn{1}{l|}{} &
  \multicolumn{3}{c|}{Recall\_subset@K} &
  \multicolumn{1}{c}{} \\ \cline{4-6}
\multicolumn{1}{c|}{\multirow{-2}{*}{Method}} &
  \multicolumn{1}{c|}{\multirow{-2}{*}{settings}} &
  \multicolumn{1}{l|}{\multirow{-2}{*}{R@1}} &
  K=1 &
  K=2 &
  \multicolumn{1}{c|}{K=3} &
  \multicolumn{1}{c}{\multirow{-2}{*}{(R@5+R\_sub@1)/2}} \\ \cline{1-3} \cline{7-7} \hline\hline
Baseline      & -   & 42.62 & 68.98 & 86.73 & 94.16 &72.84   \\
Baseline & {R+$\Delta$} & 28.25 & 65.51 & 84.91 & 93.21 &62.78  \\
SSN           & -   & 45.13 & 73.04 & 88.64 & 95.17 & 75.265  \\
SSN           & {R+$\Delta$} & 26.12 & 59.08 & 81.42 & 91.60 &57.90 \\
\bottomrule[1.5pt]
\end{tabular}}
\end{center}
\caption{Sensitivity to the change of reference images. The results indicate that the proposed SSN model is more sensitive to the reference image features. This also verifies that the core information of the reference image is retained as the prototype and plays an important role in the retrieval.}
\label{table1:sensitivity_analysis}
\end{table}
\section{More Qualitative Results}
\label{appendix:qualitative}
\begin{figure*}
\begin{center}
\includegraphics[width=\linewidth]{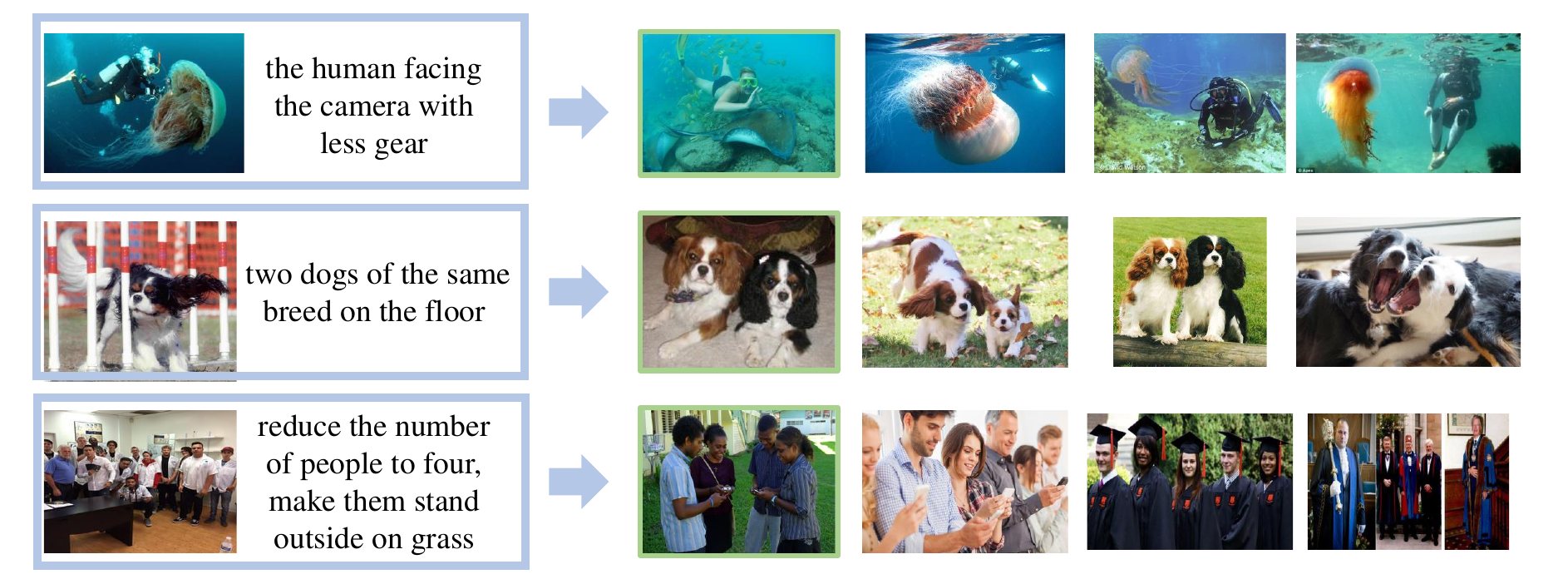}
\end{center}
   \caption{Qualitative examples of our SSN model Top-4 retrieved results on the CIRR dataset. The correct ground-truth target is denoted by a green box. The proposed SSN successfully retrieves the most relevant target images correctly.}
\label{fig:qualitative_results}
\end{figure*}

\textbf{Retrieval Examples on the CIRR Dataset.}
In this section, we show the qualitative results of our proposed model on the CIRR dataset to demonstrate the effectiveness of the SSN model.
Figure~\ref{fig:qualitative_results} gives our SSN model retrieved examples on the validation set of the CIRR dataset given a pair of reference images and text modifiers. In the first row, the background of the reference image is the ocean, and it can be observed that all the returned Top-4 images of our model are able to keep the background of the ocean unchanged. This verifies that our model can capture the core information of the reference image as visual prototypes and retain relevance between reference images and desired images in the process of search. The proposed SSN model is able to retrieve the desired target image and rank it first. For the second retrieval example, our model returns images of two dogs with the same breed on the ground as required by the text, and they both basically match the text semantic shifts. However, since the text does not state a change to the size of the dog, the second image with a dog and its puppy ranked after the first ground-truth target image. In the case of complex reference images, as the example in the third row, our model can also retrieve the target image correctly.
\begin{figure*}
\begin{center}
\includegraphics[width=\linewidth]{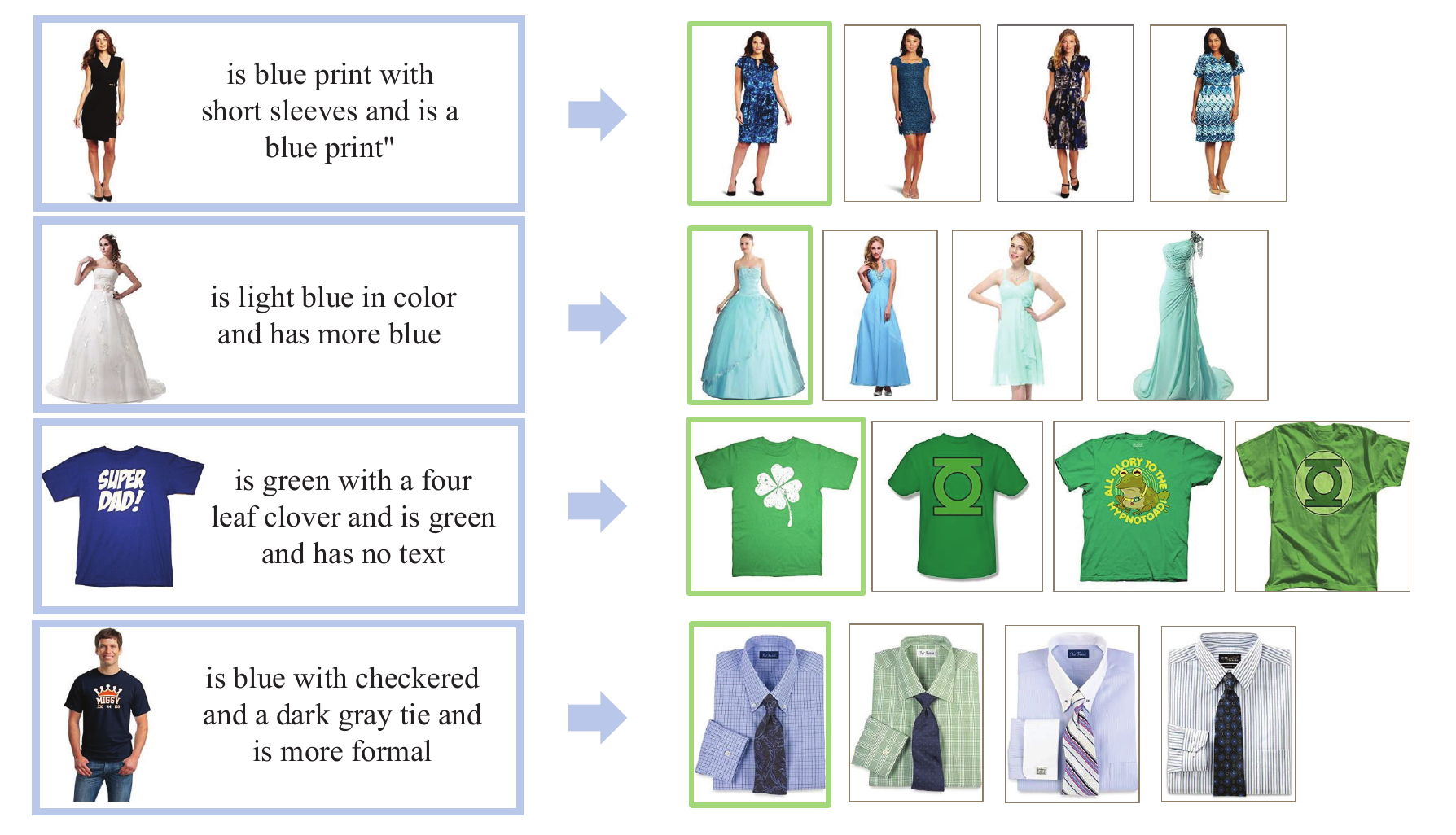}
\end{center}
   \caption{Qualitative examples of our SSN model Top-4 retrieved results on the FashionIQ dataset. The correct ground-truth target is denoted by a green box. The proposed SSN successfully retrieves the most relevant target images correctly. It can well capture the semantic shift of color change and understand some complex concepts, eg ``four-leaf clover'', ``checkered shapes''.}
\label{fig:fiq_retrieval_examples}
\end{figure*}

\noindent\textbf{Retrieval Examples on the FashionIQ Dataset}.
In Figure~\ref{fig:fiq_retrieval_examples}, we present the retrieval results of the proposed SSN model on the FashionIQ~\cite{wu2021fashion} dataset. Within the dress category (the first two rows), our model SSN well captures the color change and understands the concept of ``sleeve''. For the T-shirt category (third and fourth rows), in addition to the semantic shift of colors, our model can correctly understand some complex concepts, eg, ``four-leaf clover'', and ``checkered shapes''.

\begin{figure*}
\begin{center}
\includegraphics[width=\linewidth]{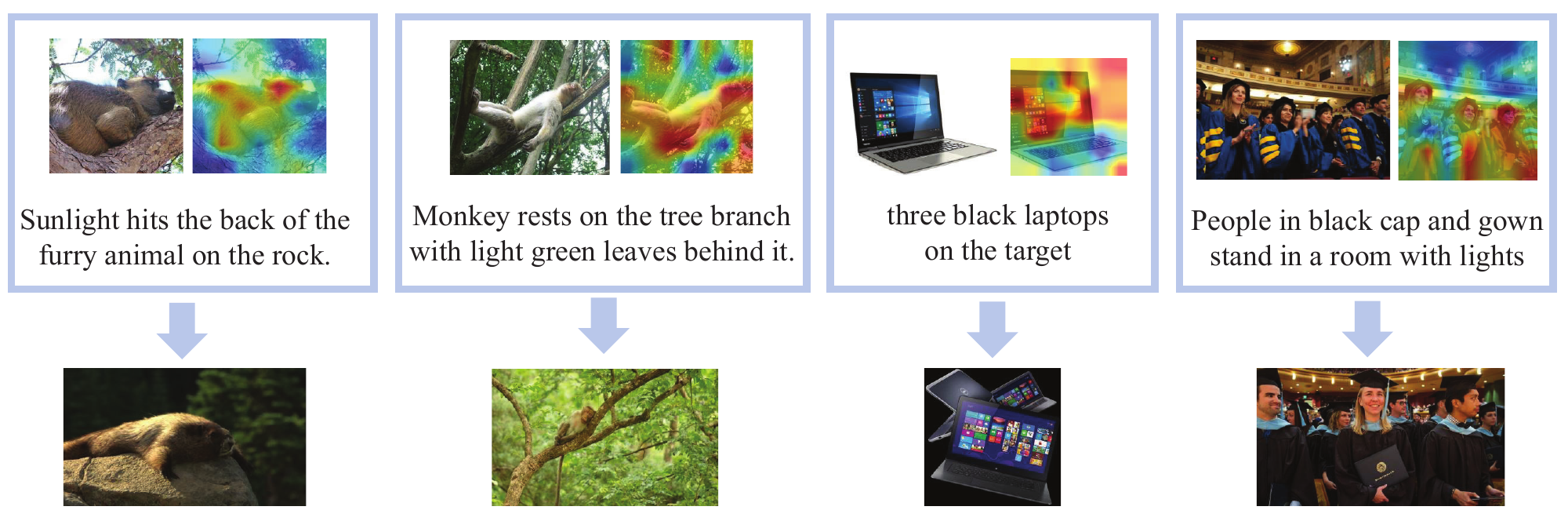}
\end{center}
   \caption{Visualization of where to be concerned when picturing visual prototypes from reference images, in the form of heatmaps. There are three examples on the validation set of the CIRR dataset, where the correct target image is ranked first within images retrieved by the given query. With decomposing semantic shifts as guidance, we observe that the majority of important information in the reference image receives attention instead of focusing on a localized characteristic. This helps to retrieve the ground-truth target image. }
\label{fig:cirr_heatmaps}
\end{figure*}

\begin{figure*}
\begin{center}
\includegraphics[width=\linewidth]{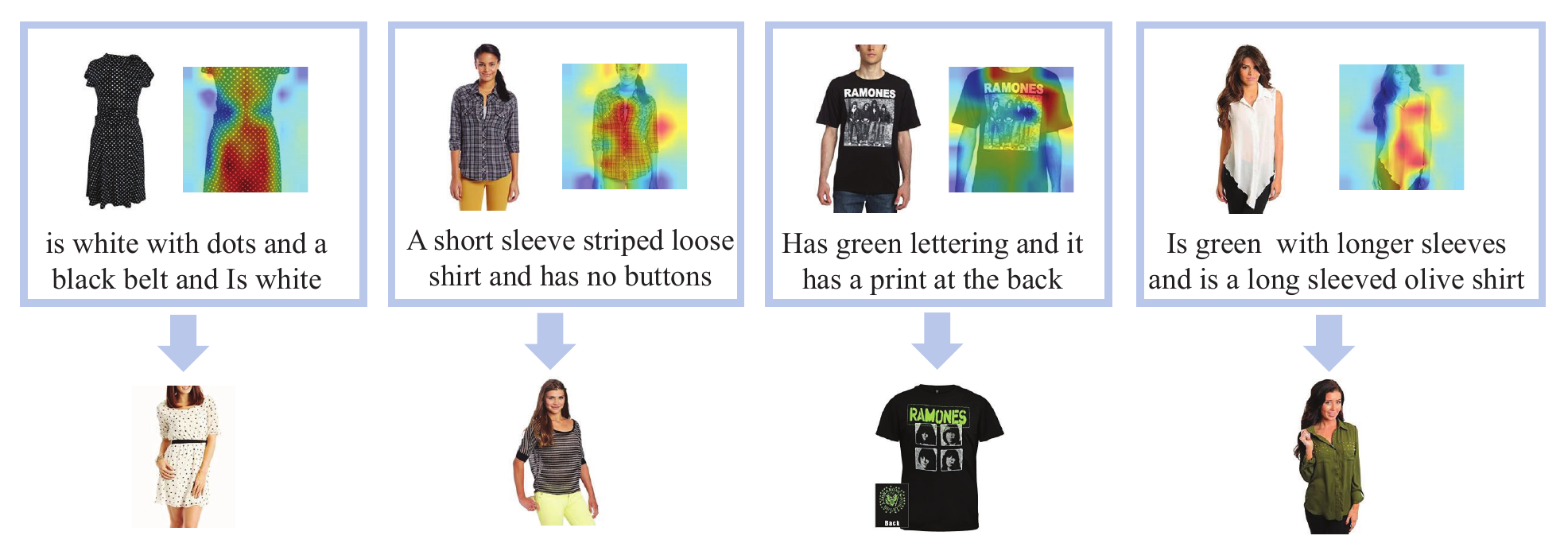}
\end{center}
   \caption{Visualization of where to be concerned when picturing visual prototypes from reference images, in the form of heatmaps. There are three examples on the validation set of the FashionIQ dataset, where the correct target image is ranked first within images retrieved by the given query. With decomposing semantic shifts as guidance, we observe that the majority of important information in the reference image receives attention instead of focusing on a localized characteristic. This helps to retrieve the ground-truth target image. }
\label{fig:fiq_heatmaps}
\end{figure*}

\noindent\textbf{Heatmaps on two datasets}.
In this section, we provide more heatmap examples of the two datasets, which shows what details are retained in the process of picturing visual prototypes from reference images on both CIRR~\cite{DBLP:conf/iccv/0002OTG21} and FashionIQ~\cite{wu2021fashion}. We plot on the basis of $C_r$, the weight for decomposing reference image features. From the heatmaps, we can tell where the visual prototype and the original reference image have changed and the extent of these changes. All results come from examples in Figure~\ref{fig:cirr_heatmaps},\ref{fig:fiq_heatmaps} that are correctly retrieved by the composed query. With decomposing semantic shifts as guidance, we observe that the majority of important information in the reference image receives attention instead of focusing on a localized characteristic. 

More specifically, in the reference image of the first group in Figure~\ref{fig:cirr_heatmaps}, the back of the furry animal is highlighted in the visual prototype, indicating the main characteristics in the target image. For the reference image of the second group, the background information is also of concern because of the ``leaves behind'' in the text. This suggests that we are not just concerned with the salient objects in the image and that the visual prototype contains a rich set of visual cues. 

Similar conclusions can be found in Figure~\ref{fig:fiq_heatmaps}. In the reference image of the first group, the dotted pattern on the dress is preserved to retrieve the correct target image. For the reference image in the second group, the visual prototype is a variation on the components of the reference image such as buttons, sleeves, and the whole pattern based on the different attention. In the reference image of the last group, the reference image changes the print (lettering and color) in the front of the T-shirt to bring it back to the visual prototype.
\end{document}